\documentclass[conference]{IEEEtran}
\IEEEoverridecommandlockouts
\usepackage{cite}
\usepackage{amsmath,amssymb,amsfonts}
\usepackage{algorithmic}
\usepackage{graphicx}
\usepackage{textcomp}
\usepackage{xcolor}
\usepackage{float}
\usepackage{rotating}
\usepackage{ upgreek }
\usepackage{mathtools}
\usepackage{diagbox}

\usepackage{graphicx}
\usepackage[font={sf,scriptsize},           
            labelfont={bf,color=accessblue},
            caption=false]                  
           {subfig}

\usepackage{comment}

\newsavebox\mybox

\def\BibTeX{{\rm B\kern-.05em{\sc i\kern-.025em b}\kern-.08em
    T\kern-.1667em\lower.7ex\hbox{E}\kern-.125emX}}
\begin{document}

\title{
Wavelet-Based Hybrid Machine Learning Model for Out-of-distribution Internet Traffic Prediction\\
}

\author{\IEEEauthorblockN{Sajal Saha, Anwar Haque, and\ *Greg Sidebottom}
\IEEEauthorblockA{\textit{Department of Computer Science} \\
\textit{University of Western Ontario, London, ON, Canada}\\
\textit{*Juniper Networks, Kanata, ON, Canada}\\
Email:\{ssaha59, ahaque32\}@uwo.ca, *gsidebot@juniper.net}
}

\maketitle

\begin{abstract}
Efficient prediction of internet traffic is essential for ensuring proactive management of computer networks. Nowadays, machine learning approaches show promising performance in modeling real-world complex traffic. However, most existing works assumed that model training and evaluation data came from identical distribution. But in practice, there is a high probability that the model will deal with data from a slightly or entirely unknown distribution in the deployment phase. This paper investigated and evaluated machine learning performances using eXtreme Gradient Boosting, Light Gradient Boosting Machine, Stochastic Gradient Descent, Gradient Boosting Regressor, CatBoost Regressor, and their stacked ensemble model using data from both identical and out-of distribution. Also, we proposed a hybrid machine learning model integrating wavelet decomposition for improving out-of-distribution prediction as standalone models were unable to generalize very well. Our experimental results show the best performance of the standalone ensemble model with an accuracy of 96.4\%, while the hybrid ensemble model improved it by 1\% for in-distribution data. But its performance dropped significantly when tested with three different datasets having a distribution shift than the training set. However, our proposed hybrid model considerably reduces the performance gap between identical and out-of-distribution evaluation compared with the standalone model, indicating the decomposition technique's effectiveness in the case of out-of-distribution generalization.

\end{abstract}


\section{Introduction}
The internet is continuing to change how we connect with others, organize the flow of things, and communicate information across the world. The demand for network traffic has risen significantly around the globe as network technology has advanced and digital activities such as video streaming, remote conferencing, online gaming, and e-commerce has increased. However, predicting network traffic based on the historical trends is indispensable for better Quality of Service (QoS), Quality of Experience (QoE), dynamic bandwidth reservation and allocation, congestion control, admission control \cite{2}, and privacy-preserving routing \cite{3}. Several research works have been done for efficient and accurate traffic prediction based on conventional statistical models\cite{iwcmc} and modern machine or deep learning techniques. 

Real internet traffic is a non-linear time series \cite{11}, and it is challenging to develop an accurate prediction model due to time-variability, long-term correlation, self-similarity, suddenness, and chaos \cite{12} in the internet traffic. Despite those non-linear characteristics, machine learning and deep learning-based methods have impactful performance. Most of the works assumed data is independent and identically distributed (i.i.d), i.e., the train and test data for model development and evaluation came from similar distribution. But in practice, the data distribution will not always be the same due to heterogeneous and anomalous internet traffic. In addition, there is a rising indication that deep learning and machine learning models exploit undesired results due to selection biases, confounding factors, and other biases in the data \cite{99}. These biases in the datasets failed in generalizing the prediction rules for the out-of-distribution data \cite{95}. Also, they assist predictive models in minimizing empirical risk by relying on correlation rather than causality. As a result, it is concerning for real-world AI (Artificial Intelligence) solutions in sensitive domains such as self-driving cars \cite{7}, health care \cite{79}, etc. 

Efficient traffic prediction is crucial for effective business decisions such as infrastructure expansion or abridgement, new service adaption, etc. Machine learning models \cite{isncc} would be an excellent solution for handling the dynamic nature of the internet traffic and providing better predictions. However, the traffic prediction tool is more likely to deal with a data distribution that is slightly different and utterly unknown than the distribution used for model training and testing. Therefore, it is essential to build a robust machine learning model that will be able to handle a shift in the data distribution. This work proposes a hybrid machine learning model based on discrete wavelet transformation to improve the out-of-distribution performance. The main contributions of our work are as follows:
\begin{itemize}
 \item Proposing a hybrid machine learning model architecture to analyze and predict real-world internet traffic.
\item Investigating different regressor models and their ensemble for single-step traffic prediction.
\item Evaluating of the model performance based on different feature lengths for identifying the optimum window length. 
\item Comparing the performance among standalone and hybrid models using four different traffic datasets to show the effectiveness of our proposed model in case of out-of-distribution generalization. 
\end{itemize}

This paper is organized as follows. Section \ref{Literature Review} describes the literature review of  current traffic prediction using machine learning models. Section  \ref{Methodology} presents the methodology, including dataset description, deep learning models explanation, anomaly identification process, and experiment details. Section  \ref{Results and Discussion} summarizes the different deep learning methods' performance and draws a comparative picture among prediction models with and without outliers in the dataset. Finally, section  \ref{Conclusion} concludes our paper and sheds light on future research directions.

\section{Literature Review}
\label{Literature Review}



A comparative performance analysis between a conventional statistical model,  ARIMA, and a deep learning model, LSTM, has been conducted in \cite{l5} for internet traffic prediction. In addition, they used the signal decomposition technique, discrete wavelet transforms (DWT), to separate the linear and non-linear components from the original data before feeding them into the prediction model. The performance of different deep sequence model is investigated in \cite{icc}. They treated anomaly in the time series data before feeding into prediction model and showed the effectiveness of outlier detection in internet traffic prediction. A hybrid model consisting of a statistical and deep learning model is proposed in \cite{l1} for better performance than the standalone model. In addition, they applied Discrete Wavelet Transform (DWT) on the time series data for separating the linear and non-linear components, modeled respectively using Auto Regressive Moving Average (ARIMA) and Recurrent Neural Networks (RNN). Since the conventional statistical models such as ARMA, and ARIMA, are incapable of handling the non-linear components in the time series, the author tried to use the signal decomposition technique here to deal with the complex, non-stationary internet traffic by combining the power of deep learning. According to their experimental results, the combination of ARIMA and RNN performed better than the individual model. 

An artificial neural network model combined with the multi-fractal DWT is proposed in \cite{l2}. The network traffic is discomposed into low frequency and high-frequency components using Haar Wavelet, which has been considered a target for the ANN model with the input of the original traffic data. In the end, model predictions are combined to reconstruct the actual value. Their model outperformed compared with two existing methodologies. In \cite{l4}, they performed a comparative analysis among different methods of DWT and spline-extrapolation in predicting the characteristics of the IoT multimedia internet traffic. The spline-extrapolation with the B-splines was the best, giving them the minimum forecast error of 5\% compared with Haar-wavelet and quadratic splines having prediction errors respectively 7-10\% and 10\%. In \cite{l3}, the author developed a traffic prediction framework by combining the power of several technical methods such as Mallat wavelet transformation, Hurst exponent analysis, model parameter optimization, and fusion of multiple prediction models. Firstly, a three-level decomposition has been carried out on the original traffic data to extract a set of approximate and detailed components. Then, the individual component predictability was analyzed using Hurst exponent analysis, where a higher Hurst exponent H indicates more predictableness. According to their study, the approximate component has a higher H than the detailed component. As a result, a more efficient machine learning model, the Least squares Support vector machine (LSSVM), is used to predict detail components while the approximate component is analyzed using ARIMA. The proposed method showed better prediction accuracy compared with the other models. 

The use of signal decomposition in real-world internet traffic prediction is prevalent in the current literature. The existing works indicate the outperformance of the hybrid model capable of handling the linear and non-linear components separately using different types of models. However, most research assumes that the model’s data come from a similar distribution, although the scenario is quite different in practice. The prediction model is more likely to see a distribution shift or completely unknown distribution when it will deploy in the production. According to recent studies, the machine learning methods do not guarantee the generalization of the out-of-distribution data. And this issue is not considered extensively in the existing literature. Therefore, it is significantly vital to build a robust prediction model so that it can generalize the unknown distribution in the future. In this research, we propose a hybrid machine learning model that can better generalize the distribution shift in the data rather than the individual model.   
\begin{figure}
    \centering
    \includegraphics[height=5cm,width=9cm]{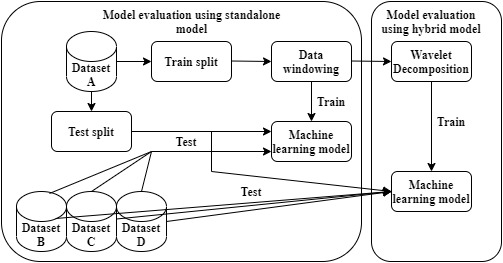}
    \caption{High-level methodology}
    \label{fig:methodology}
\end{figure}
\section{Methodology}
\label{Methodology}

\subsection{Dataset and Preprocessing Steps}
\label{data_preprocessing}
Real internet traffic telemetry on several high-speed interfaces has been used for this experiment. The data are collected every five minutes for a recent thirty days time period. A total of five different datasets are used in our experiment. The source domain dataset, Dataset A, consists of 8563 data samples and it is used to build the predictive model for the source task. The other four datasets, Dataset B, C, and D, and E are comparatively smaller in size, having 363, 369, and 358 data instances, respectively. Time series data need to be expressed in the proper format for supervised learning. Generally, the time-series data consists of several tuples (time, value), which is inappropriate for feeding them into the machine learning model. So, we restructured our original time series data using the sliding window technique. The high-level diagram of our methodology is shown in Fig.\ref{fig:methodology}.

\subsection{Discrete Wavelet Transform}
\label{dwt}
The wavelet transform is a mathematical way of finding the hidden patterns in the original dataset by decomposing the signal into a time-frequency domain. The process involves a wavelet, i.e., a wave-like oscillation for extracting multiple lower resolution levels by controlling the scale and location of the wavelet \cite{dt1}. There are two main types of wavelet transform, Discrete Wavelet Transform ($DWT$) and Continuous Wavelet Transform ($CWT$). In this work, we used $DWT$ for the decomposition step because $CWT$ will be a very time-consuming transform with extra and useless data. At each stage of $DWT$, the signal is decomposed into two components: approximate component, $Ca$ and detailed component, $Cd$ representing the general trend and detailed events in the data, respectively. The $Ca$  from level $i$ is used to calculate the $Ca$ in the next level, $i+1$. A low-pass filter ($l_p$) and high-pass filter ($h_p$) convolute the signal to generate the new $Ca_{i+1}$ and $Cd_{i+1}$. The equation for components is as follows \cite{dt2}:
\begin{equation}
   Ca_{i+1}[n] = Ca_{i}*l_p[n] = \sum_{m=-\infty}^{\infty} Ca_i[m]l_p[n-2m]
\end{equation}   
\begin{equation}
   Cd_{i+1}[n] = Ca_{i}*h_p[n] = \sum_{m=-\infty}^{\infty} Ca_i[m]h_p[n-2m]
\end{equation}
Here $n$ is the total number of levels. We can reconstruct the original data by combining all level's detailed components  and approximate components from the last level.

\begin{equation}
    \text{Original data, } y_t = Ca_n + \sum_{i=1}^{n}Cd_i
\end{equation}

\begin{table*}
\centering
\caption{Performance of all models train and test on dataset A}
\label{tab:prediction_summary}
\begin{tabular}{|c|c|c|c|c||c|c|c|c||c|c|c|c||c|c|c|c|} 
\hline
    & \multicolumn{4}{c||}{\textbf{No Wavelet}} & \multicolumn{4}{c||}{\textbf{dmey}} & \multicolumn{4}{c||}{\textbf{haar}} & \multicolumn{4}{c|}{\textbf{bior3.7}}  \\ 
\hline
    & 6    & 9    & 12   & 15                   & 6    & 9    & 12   & 15             & 6    & 9    & 12   & 15             & 6    & 9    & 12   & 15                \\ 
\hline
XGB & 96.1 & 96.0 & 95.9 & 95.9                 & 97.4 & 97.4 & 97.4 & 97.3           & 97.2 & 97.3 & 97.3 & 97.2           & 97.2 & 97.2 & 97.2 & 97.3              \\ 
\hline
LGB & 96.2 & 96.2 & 96.1 & 96.2                 & 97.4 & 97.4 & 97.4 & 97.4           & 97.3 & 97.3 & 97.3 & 97.3           & 97.1 & 97.2 & 97.2 & 97.2              \\ 
\hline
SGD & 96.0 & 96.1 & 96.2 & 96.1                 & 96.7 & 96.9 & 97.0 & 97.0           & 96.7 & 96.6 & 96.9 & 96.9           & 96.7 & 96.8 & 96.8 & 96.9              \\ 
\hline
GBR & 96.1 & 96.1 & 96.2 & 96.2                 & 97.1 & 97.1 & 97.1 & 97.2           & 97.1 & 97.1 & 97.1 & 97.1           & 97.1 & 97.2 & 97.2 & 97.2              \\ 
\hline
CAB & 96.3 & 96.1 & 96.2 & 96.2                 & 97.3 & 97.2 & 97.2 & 97.2           & 97.2 & 97.1 & 97.1 & 97.1           & 97.2 & 97.2 & 97.2 & 97.1              \\ 
\hline
ENS & 96.1 & 96.3 & 96.3 & 96.4                 & 97.4 & 97.4 & 97.3 & 97.3           & 97.2 & 97.2 & 97.1 & 97.2           & 97.1 & 97.1 & 97.1 & 97.0              \\
\hline
\end{tabular}
\end{table*}
\begin{table*}
\centering
\caption{Performance of all models train on dataset A, test on dataset B}
\label{tab:datasets b}
\begin{tabular}{|c|c|c|c|c||c|c|c|c||c|c|c|c||c|c|c|c|} 
\hline
    & \multicolumn{4}{c||}{\textbf{No Wavelet}}           & \multicolumn{4}{c||}{\textbf{dmey}}                 & \multicolumn{4}{c||}{\textbf{haar}}                 & \multicolumn{4}{c|}{\textbf{bior3.7}}                \\ 
\hline
    & \textbf{6} & \textbf{9} & \textbf{12} & \textbf{15} & \textbf{6} & \textbf{9} & \textbf{12} & \textbf{15} & \textbf{6} & \textbf{9} & \textbf{12} & \textbf{15} & \textbf{6} & \textbf{9} & \textbf{12} & \textbf{15}  \\ 
\hline
XGB & 78.9       & 78.7       & 80.1        & 78.0        & 87.9       & 87.6       & 87.4        & 87.4        & 87.3       & 87.5       & 87.5        & 87.1        & 83.3       & 81.9       & 82.1        & 82.8         \\ 
\hline
LGB & 83.4       & 83.3       & 82.9        & 82.8        & 87.6       & 87.3       & 87.1        & 86.9        & 87.3       & 87.4       & 87.2        & 87.0        & 84.8       & 84.5       & 84.1        & 83.9         \\ 
\hline
SGD & 80.1       & 79.5       & 78.6        & 77.9        & 84.1       & 83.8       & 82.8        & 82.6        & 84.0       & 83.3       & 82.5        & 81.7        & 83.6       & 83.3       & 82.3        & 81.4         \\ 
\hline
GBR & 82.6       & 81.6       & 82.0        & 81.0        & 86.7       & 86.9       & 86.8        & 86.7        & 86.5       & 86.2       & 85.6        & 85.5        & 82.5       & 82.5       & 82.0        & 81.5         \\ 
\hline
CAB & 80.6       & 79.8       & 81.3        & 80.4        & 85.7       & 84.9       & 84.8        & 84.0        & 85.6       & 85.2       & 85.3        & 84.8        & 83.8       & 83.2       & 83.0        & 81.3         \\ 
\hline
ENS & 79.3       & 80.3       & 80.5        & 79.9        & 86.9       & 85.5       & 85.4        & 84.9        & 86.0       & 85.5       & 86.0        & 85.2        & 84.9       & 84.4       & 83.8        & 83.4         \\
\hline
\end{tabular}
\end{table*}
\begin{table*}
\centering
\caption{Performance of all models train on dataset A, test on dataset C}
\label{tab:datasets c}
\begin{tabular}{|c|c|c|c|c||c|c|c|c||c|c|c|c||c|c|c|c|} 
\hline
    & \multicolumn{4}{c||}{\textbf{No Wavelet}}           & \multicolumn{4}{c||}{\textbf{dmey}}                 & \multicolumn{4}{c||}{\textbf{haar}}                 & \multicolumn{4}{c|}{\textbf{bior3.7}}                \\ 
\hline
    & \textbf{6} & \textbf{9} & \textbf{12} & \textbf{15} & \textbf{6} & \textbf{9} & \textbf{12} & \textbf{15} & \textbf{6} & \textbf{9} & \textbf{12} & \textbf{15} & \textbf{6} & \textbf{9} & \textbf{12} & \textbf{15}  \\ 
\hline
XGB & 72.6       & 71.7       & 72.9        & 71.0        & 84.0       & 83.6       & 83.2        & 83.1        & 82.2       & 82.5       & 82.3        & 82.3        & 80.7       & 79.7       & 79.6        & 80.2         \\ 
\hline
LGB & 75.6       & 75.9       & 75.4        & 75.1        & 83.5       & 83.3       & 83.2        & 83.1        & 82.2       & 82.2       & 82.0        & 81.9        & 82.0       & 81.7       & 81.4        & 81.1         \\ 
\hline
SGD & 73.9       & 73.6       & 73.3        & 72.6        & 79.7       & 79.5       & 78.9        & 78.9        & 78.6       & 78.4       & 77.9        & 77.3        & 79.6       & 79.2       & 78.9        & 78.1         \\ 
\hline
GBR & 74.1       & 72.8       & 73.9        & 73.0        & 83.1       & 83.1       & 82.6        & 82.8        & 80.9       & 81.1       & 80.8        & 80.6        & 79.1       & 78.7       & 78.7        & 78.9         \\ 
\hline
CAB & 74.0       & 73.2       & 73.6        & 73.8        & 82.5       & 81.4       & 81.8        & 81.1        & 80.8       & 80.6       & 80.4        & 80.0        & 80.9       & 80.9       & 81.1        & 80.0         \\ 
\hline
ENS & 73.5       & 74.4       & 74.3        & 73.7        & 82.6       & 81.2       & 80.7        & 80.9        & 80.8       & 80.6       & 80.5        & 80.0        & 81.4       & 81.2       & 80.3        & 80.4         \\
\hline
\end{tabular}
\end{table*}
\begin{table*}
\centering
\caption{Performance of all models train on dataset A, test on dataset D}
\label{tab:datasets d}
\begin{tabular}{|c|c|c|c|c||c|c|c|c||c|c|c|c||c|c|c|c|} 
\hline
    & \multicolumn{4}{c||}{\textbf{No Wavelet}}           & \multicolumn{4}{c||}{\textbf{dmey}}                 & \multicolumn{4}{c||}{\textbf{haar}}                 & \multicolumn{4}{c|}{\textbf{bior3.7}}                \\ 
\hline
    & \textbf{6} & \textbf{9} & \textbf{12} & \textbf{15} & \textbf{6} & \textbf{9} & \textbf{12} & \textbf{15} & \textbf{6} & \textbf{9} & \textbf{12} & \textbf{15} & \textbf{6} & \textbf{9} & \textbf{12} & \textbf{15}  \\ 
\hline
XGB & 85.3       & 84.9       & 85.1        & 85.0        & 89.4       & 89.4       & 89.1        & 89.4        & 89.1       & 89.1       & 89.1        & 89.1        & 87.8       & 87.2       & 87.2        & 87.4         \\ 
\hline
LGB & 86.6       & 86.5       & 86.4        & 86.6        & 89.6       & 89.5       & 89.4        & 89.6        & 89.3       & 89.2       & 89.1        & 89.3        & 89.8       & 89.5       & 89.3        & 89.4         \\ 
\hline
SGD & 85.0       & 84.5       & 84.5        & 84.7        & 87.5       & 87.2       & 87.0        & 87.1        & 87.4       & 86.9       & 86.9        & 87.0        & 88.2       & 87.7       & 87.6        & 87.7         \\ 
\hline
GBR & 86.5       & 86.0       & 86.3        & 86.2        & 88.6       & 88.7       & 88.7        & 88.7        & 88.7       & 88.6       & 88.5        & 88.8        & 87.0       & 87.0       & 86.9        & 87.1         \\ 
\hline
CAB & 86.3       & 86.2       & 86.0        & 86.2        & 89.6       & 89.5       & 89.6        & 89.7        & 89.3       & 89.4       & 89.2        & 89.3        & 89.5       & 89.1       & 88.7        & 88.6         \\ 
\hline
ENS & 86.1       & 85.9       & 86.0        & 86.1        & 89.2       & 88.6       & 88.6        & 88.3        & 88.6       & 88.4       & 88.2        & 88.2        & 89.2       & 89.1       & 88.5        & 89.1         \\
\hline
\end{tabular}
\end{table*}

\subsection{Out-of-distribution Generalization}
\label{ood}
Machine learning model dealt with a feature set $X$ and corresponding target $Y$ that could be a discrete value (classification task) or continuous value (regression task). The purpose of the machine learning algorithm is to optimize the learnable parameters theta for a function $f: X \longrightarrow Y$. The function parameter theta is optimized based on the loss function $L:Y \longrightarrow Y$ which returns the gap between actual and predicted value, and the purpose of the machine learning model is to find the function parameters theta with minimum loss. Many machine learning models have been developed based on the assumption that the train data and test data are come from same distribution i.e. $P_{tr} (X,Y)$ = $P_{te}(X,Y)$. But, the train and test distribution can be different due to several reasons such as temporal/spatial evolution of data or the sample selection bias in data collection process. Therefore, out-of-distribution generalization discusses machine learning methodology where a distribution shift is exists between $P_{tr}$ and $P_{te}$. In OOD problem settings, we need to define how the test distribution is different from the train distribution. There are different distribution shifts in the literature but the most common one is the covariate shift where the target generation process is the same with the marginal distribution of $X$ shifts from the training set to the test set. In this work, we deal with the covariate shifts in the data.


\subsection{Evaluation Metrics}
\label{metric}
We used Weighted Average Percentage Error (WAPE) to estimate the performance of our traffic forecasting models. The performance metric identifies the deviation of the predicted result from the original data. For example, WAPE error represents the average percentage of fluctuation between the actual value and predicted value. Therefore, we can define our performance metric mathematically as follow: 
\begin{equation}
    WAPE =\dfrac {\sum_{i=1}^n |{p_i-o_i}|} {\sum_{i=1}^n |{o_i}|} \times 100 \%
\end{equation}

\begin{equation}
    Accuracy = (100-WAPE)\% 
\end{equation}

Here, $p_i$ and $o_i$ are predicted and original value respectively and  $n$ is the total number of test instance.

\subsection{Software and Hardware Preliminaries}
\label{software}
We used Python and machine learning library Scikit-Learn\cite{scikit-learn} and PyWavelets\cite{lee2019pywavelets} to conduct the experiments.  Our computer has the configuration of Intel (R) i3-8130U CPU@2.20GHz, 8GB memory, and a 64-bit Windows operating system. 

\begin{figure}
    \centering
    \includegraphics[height=3.5cm,width=9cm]{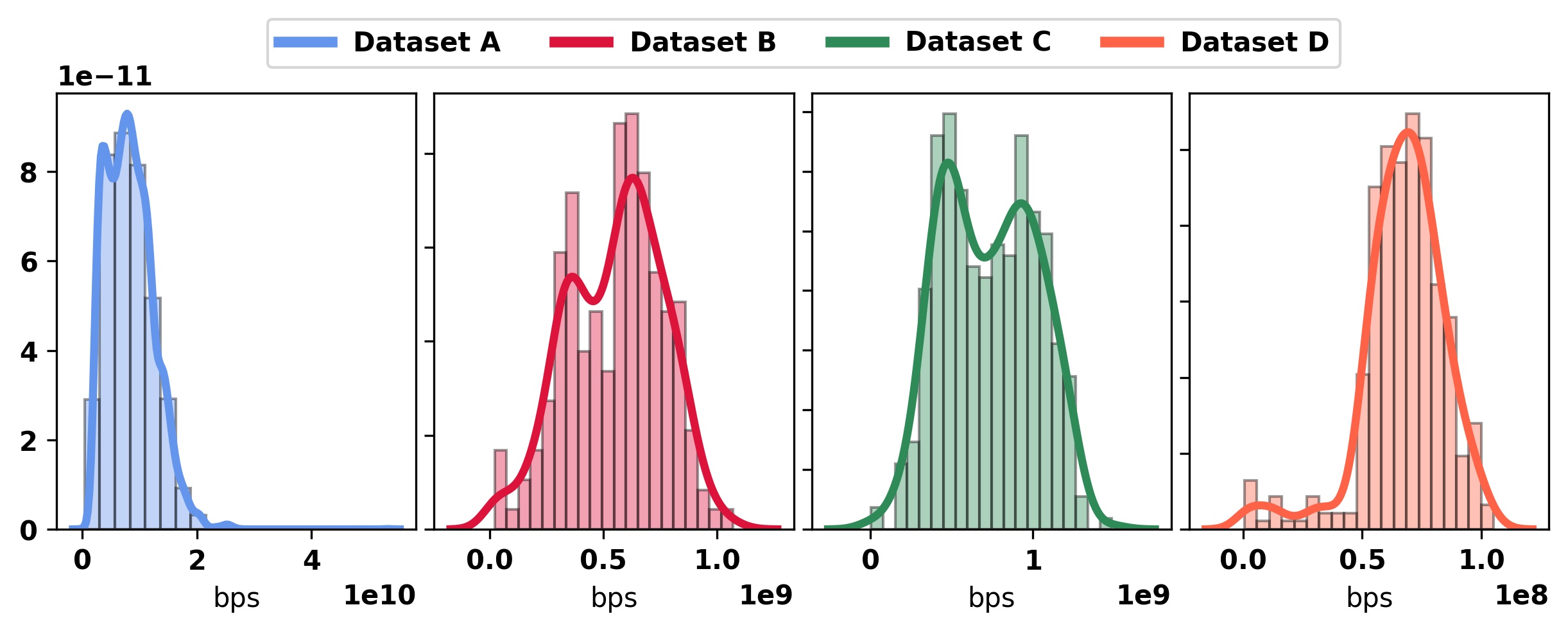}
    \caption{Data distribution for different datasets used in the experiment}
    \label{fig:density plot}
\end{figure}

\section{Results and Discussion}
\label{Results and Discussion}
\subsection{Experimental setup}
\label{sss:experimental setup}
As we mentioned earlier, our main research objectives are to improve the machine learning model's performance in case of a distribution shift. We used four different internet traffic datasets extracted from real-world telemetry to conduct this experiment. But all the datasets are not equivalent according to the data volume, and their distribution is different, as represented in Fig.\ref{fig:density plot}. Therefore, we used a comparatively larger dataset $A$ to build our traffic prediction model. The other datasets from $B$ to $D$ are the test datasets with distribution shifts. So, our experiment has three main parts: i) design and validate traffic prediction model based on dataset $A$ ii) validate the model performance with other three datasets $B$, $C$, $D$, which are entirely unknown to the model, and iii) finally modify models by adapting discrete wavelet transformation to improve inference on datasets $B$, $C$, $D$.

\begin{figure}
    \centering
    \includegraphics[height=4cm,width=9cm]{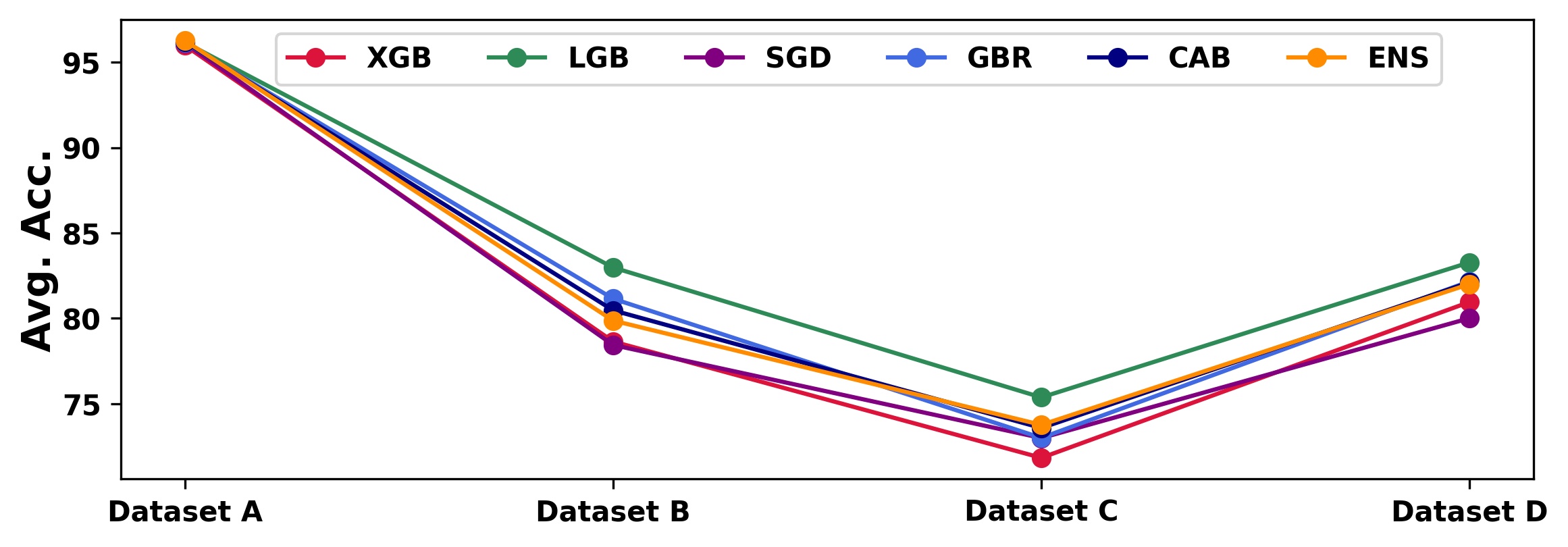}
    \caption{Average accuracy of standalone model for different test set}
    \label{fig:avg_acc_standalone}
\end{figure}

\begin{figure*}
    \centering
    \includegraphics[height=3cm,width=17cm]{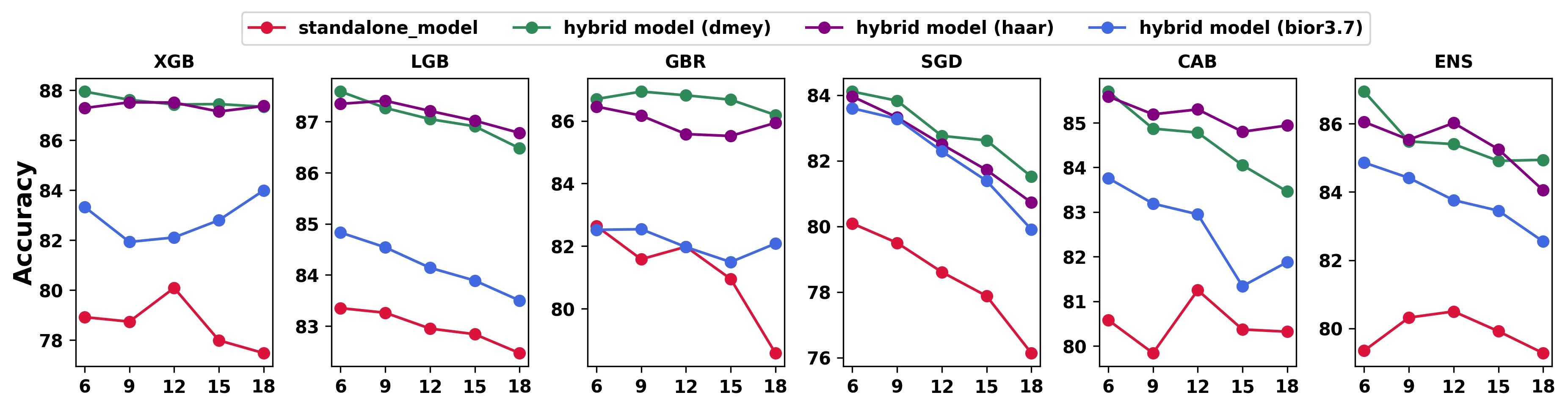}
    \caption{Performance comparison among standalone and hybrid models training on dataset A test on dataset B}
    \label{fig:dataset b}
\end{figure*}

\begin{figure*}
    \centering
    \includegraphics[height=3cm,width=17cm]{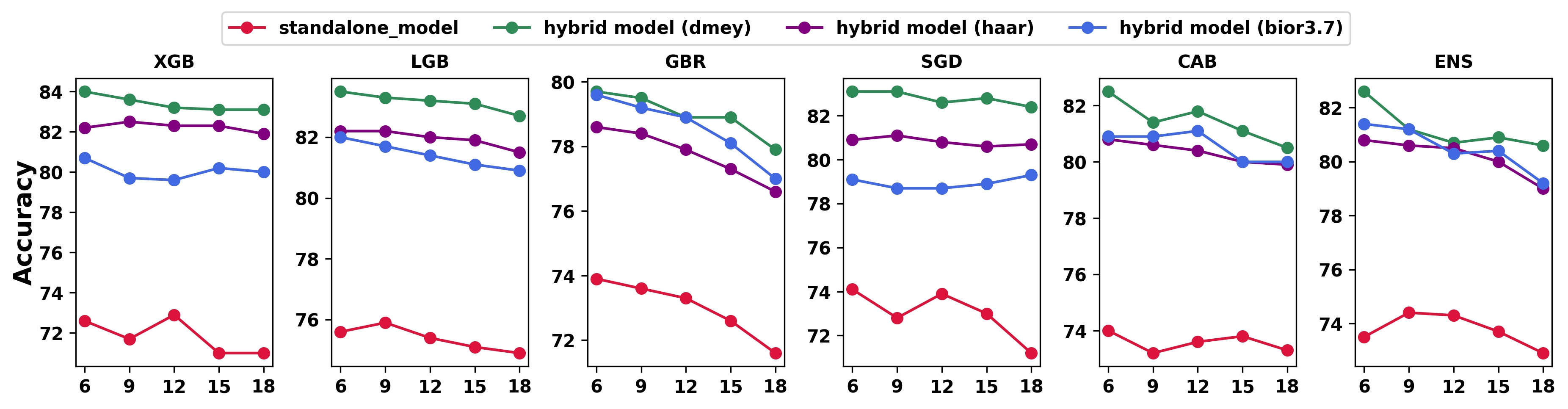}
    \caption{Performance comparison among standalone and hybrid models training on dataset A test on dataset C}
    \label{fig:dataset c}
\end{figure*}

\begin{figure*}
    \centering
    \includegraphics[height=3cm,width=17cm]{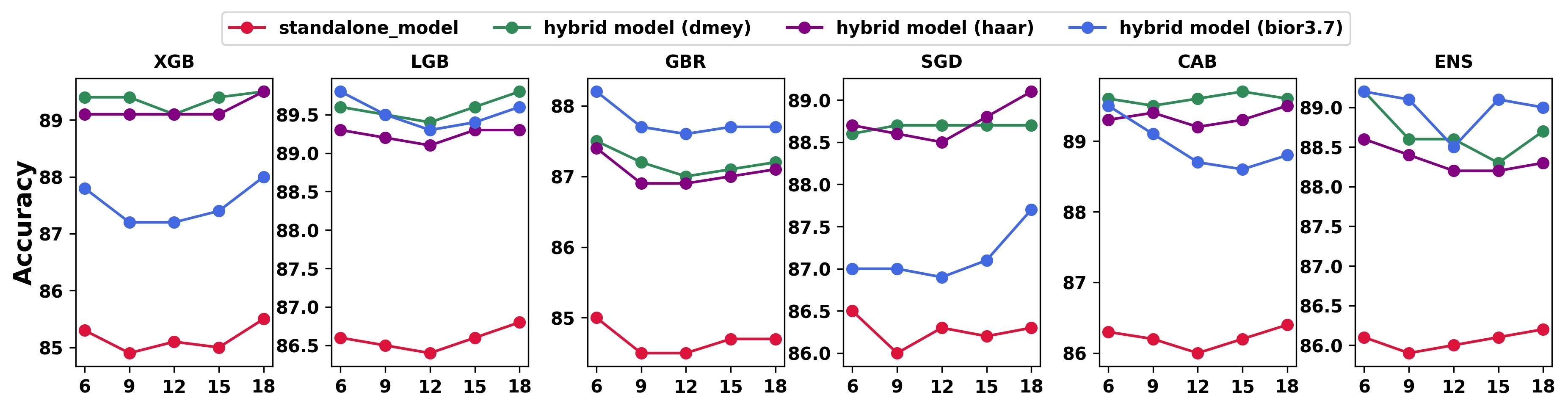}
    \caption{Performance comparison among standalone and hybrid models training on dataset A test on dataset D}
    \label{fig:dataset d}
\end{figure*}

\begin{figure*}
    \centering
    \includegraphics[height=4cm,width=17cm]{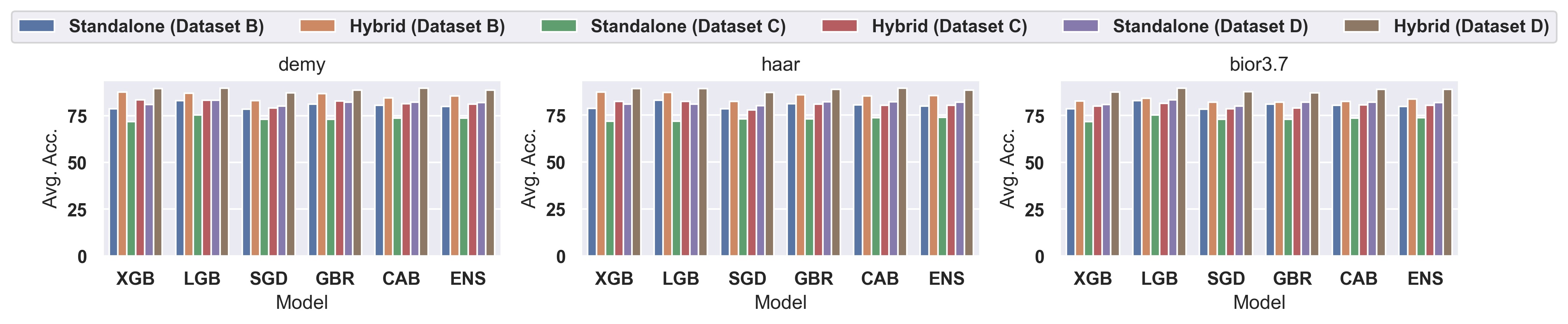}
    \caption{Standalone and hybrid model's accuracy comparison for different datasets}
    \label{fig:standalone and hybrid performance comparision}
\end{figure*}

\subsection{Result analysis}
\label{sss:result}
We applied several machine learning models such as eXtreme Gradient Boosting (XGB), Light Gradient Boosting Machine (LGB), CatBoost Regressor (CAB), Stochastic Gradient Descent (SGD), Gradient Boosting Regressor (GBR), and an ensemble model (ENS) based on the stacking technique. The performance of all machine learning models trained and validated using dataset $A$ is summarized in Table \ref{tab:prediction_summary}. The group column \lq No Wavelet\rq\text{ }represents the standalone model performance while the other three columns \lq dmey\rq,\text{ } \lq haar\rq, and \lq bior3.7\rq\text{ } represent the hybrid model with the corresponding mother wavelet. We used five different feature window lengths of 6, 9, 12, and 15 to train and identify an optimum number of inputs for the model. In the case of standalone model evaluation, the XGBoost model achieved the best prediction accuracy of 96.1\% using the six input features. The highest accuracy for LGB, SGD, GBR, and CAB are respectively 96.2\%, 96.2\%, 96.2\%, and 96.3\% using input 6, 12, 12, and 6. Finally, the ensemble model gave us the best prediction accuracy of 96.4\% using 15 input features. Then, the standalone models are modified by enabling them to accept decomposed components of the series by discrete wavelet decomposition varied by three different mother wavelets. Overall, the model performance is improved in the case of the hybrid model for each setting. However, the best hybrid model performance we achieved is 97.4\% using the ensemble model with dmey wavelet, which is a 1\% improvement over the standalone model. 
After evaluating models with a test set from the same training set distribution, we used three other datasets from different distributions to validate model performance in case of out-of distribution. The results are summarized in Table \ref{tab:datasets b}, Table \ref{tab:datasets c}, and Table \ref{tab:datasets d} respectively for dataset $B$, dataset $C$, and dataset $D$. As the distribution of these three datasets differs from the dataset used in training, the model performance decreases significantly. Fig. \ref{fig:avg_acc_standalone} depicts the performance trend of different prediction model for four different test distribution.  Overall, the model performance was at its peak when it was evaluated using the same distribution data, i.e., training and test on dataset $A$. The performance dropped from 95\%-96\% to 80\%-85\% range when tested using dataset $B$ having distribution shift. And the similar pattern is clear from the figure for the other two datasets. 
Finally, we evaluate our hybrid model's performance in our-of-distribution generalization. Fig \ref{fig:dataset b}, \ref{fig:dataset c}, and  \ref{fig:dataset d} represent comparative performance among standalone model and three different hybrid models. These figures help compare the performance of different mother wavelets and input sizes. Our experiment showed a significant performance enhancement after applying wavelet transformation. For example, the XGB model with dmey wavelet gave us more than 8\% accurate result for dataset $B$ compared with the standalone model.
Similarly, more than 10\% accuracy has been improved in the case of dataset $C$ by the LGB model with haar wavelet. Overall, the hybrid model's performance is better than the standalone model for each input set. In addition, the hybrid model with dmey mother wavelet performance is better on average than the other two wavelets, while the bior3.7 gave us the poorer performance. Hence, choosing the proper mother wavelet is essential for better prediction results. In addition, the effect of selecting the appropriate window size is prevalent in the diagrams. Finally, we depicted the hybrid model's performance improvement compared with the standalone model for our three datasets, different from the training dataset in Fig.\ref{fig:standalone and hybrid performance comparision}. Each model performance indicates a better generalization power when combined with the decomposition technique.

\section{Conclusion}
\label{Conclusion}
In this work, we evaluate the machine learning model's performance in case of a distribution shift in the internet traffic dataset. The majority of the machine learning models for traffic prediction are assessed based on the same distribution data. But in practice, the prediction model will most likely encounter slightly or entirely different data than the dataset used for the model development. As real-world traffic is susceptible to various internal and external factors, it is highly volatile. For our experiment, we considered a total of 4 datasets. The larger one is used for the traditional machine learning model development and validation, where the train and test split was respectively 70\% and 30\%. A total of six different machine learning models were used for the traffic prediction task, while the ensemble model gave us the best prediction accuracy of 96.4\%. After integrating the decomposition technique with the model, the best performance increased by 1\% to 97.4\% by ensemble model and dmey wavelet. Then we used similar models for inferencing our other three smaller datasets from a different distribution than the training dataset. These three datasets are entirely unknown to our prediction models. Due to the distribution shift, the model performance decreased significantly, indicating a fundamental problem of machine learning solutions after deployment in the real world. But the hybrid model performance was considerably better than the standalone model in case of out-of-distribution generalization. In the future, we would like to extend this work by adapting empirical mode decomposition to avoid the problem of choosing a suitable wavelet for a better prediction. Also, we plan to experiment with the multi-step forecast as well.

\bibliographystyle{IEEEtran}
\bibliography{ref}

\begin{thebibliography}{10}
\providecommand{\url}[1]{#1}
\csname url@samestyle\endcsname
\providecommand{\newblock}{\relax}
\providecommand{\bibinfo}[2]{#2}
\providecommand{\BIBentrySTDinterwordspacing}{\spaceskip=0pt\relax}
\providecommand{\BIBentryALTinterwordstretchfactor}{4}
\providecommand{\BIBentryALTinterwordspacing}{\spaceskip=\fontdimen2\font plus
\BIBentryALTinterwordstretchfactor\fontdimen3\font minus
  \fontdimen4\font\relax}
\providecommand{\BIBforeignlanguage}[2]{{%
\expandafter\ifx\csname l@#1\endcsname\relax
\typeout{** WARNING: IEEEtran.bst: No hyphenation pattern has been}%
\typeout{** loaded for the language `#1'. Using the pattern for}%
\typeout{** the default language instead.}%
\else
\language=\csname l@#1\endcsname
\fi
#2}}
\providecommand{\BIBdecl}{\relax}
\BIBdecl

\bibitem{2}
H.~Kalbkhani, M.~G. Shayesteh, and N.~Haghighat, ``Adaptive lstar model for
  long-range variable bit rate video traffic prediction,'' \emph{IEEE
  Transactions on Multimedia}, vol.~19, no.~5, pp. 999--1014, 2016.

\bibitem{3}
G.~Han, H.~Wang, X.~Miao, L.~Liu, J.~Jiang, and Y.~Peng, ``A dynamic multipath
  scheme for protecting source-location privacy using multiple sinks in wsns
  intended for iiot,'' \emph{IEEE Transactions on Industrial Informatics},
  vol.~16, no.~8, pp. 5527--5538, 2019.

\bibitem{iwcmc}
S.~Saha, A.~Haque, and G.~Sidebottom, ``An empirical study on internet traffic
  prediction using statistical rolling model,'' 2022, (in press).

\bibitem{11}
D.~Wei, ``Network traffic prediction based on rbf neural network optimized by
  improved gravitation search algorithm,'' \emph{Neural Computing and
  Applications}, vol.~28, no.~8, pp. 2303--2312, 2017.

\bibitem{12}
Q.~Li and R.-C. Lin, ``A new approach for chaotic time series prediction using
  recurrent neural network,'' \emph{Mathematical Problems in Engineering}, vol.
  2016, 2016.

\bibitem{99}
Z.~Shen, J.~Liu, Y.~He, X.~Zhang, R.~Xu, H.~Yu, and P.~Cui, ``Towards
  out-of-distribution generalization: A survey,'' \emph{arXiv preprint
  arXiv:2108.13624}, 2021.

\bibitem{95}
B.~Sch{\"o}lkopf, F.~Locatello, S.~Bauer, N.~R. Ke, N.~Kalchbrenner, A.~Goyal,
  and Y.~Bengio, ``Toward causal representation learning,'' \emph{Proceedings
  of the IEEE}, vol. 109, no.~5, pp. 612--634, 2021.

\bibitem{7}
C.~Badue, R.~Guidolini, R.~V. Carneiro, P.~Azevedo, V.~B. Cardoso, A.~Forechi,
  L.~Jesus, R.~Berriel, T.~M. Paixao, F.~Mutz \emph{et~al.}, ``Self-driving
  cars: A survey,'' \emph{Expert Systems with Applications}, vol. 165, p.
  113816, 2021.

\bibitem{79}
A.~Rajkomar, E.~Oren, K.~Chen, A.~M. Dai, N.~Hajaj, M.~Hardt, P.~J. Liu,
  X.~Liu, J.~Marcus, M.~Sun \emph{et~al.}, ``Scalable and accurate deep
  learning with electronic health records,'' \emph{NPJ Digital Medicine},
  vol.~1, no.~1, pp. 1--10, 2018.

\bibitem{isncc}
S.~Saha, A.~Haque, and G.~Sidebottom, ``Towards an ensemble regressor model for
  anomalous isp traffic prediction,'' 2022, (in press).

\bibitem{l5}
G.~N.~N. Barbosa, M.~A. Lopez, D.~S.~V. Medeiros, and D.~M.~F. Mattos, ``An
  entropy-based hybrid mechanism for large-scale wireless network traffic
  prediction,'' in \emph{2021 International Symposium on Networks, Computers
  and Communications (ISNCC)}, 2021, pp. 1--6.

\bibitem{icc}
S.~Saha, A.~Haque, and G.~Sidebottom, ``Deep sequence modeling for anomalous
  isp traffic prediction,'' 2022, (in press).

\bibitem{l1}
R.~Madan and P.~S. Mangipudi, ``Predicting computer network traffic: a time
  series forecasting approach using dwt, arima and rnn,'' in \emph{2018
  Eleventh International Conference on Contemporary Computing (IC3)}.\hskip 1em
  plus 0.5em minus 0.4em\relax IEEE, 2018, pp. 1--5.

\bibitem{l2}
Y.~Lu, H.~Li, B.~Lu, Y.~Zhao, D.~Wang, X.~Gong, and X.~Wei, ``Network traffic
  model with multi-fractal discrete wavelet transform in power
  telecommunication access networks,'' in \emph{International Conference on
  Simulation Tools and Techniques}.\hskip 1em plus 0.5em minus 0.4em\relax
  Springer, 2019, pp. 53--62.

\bibitem{l4}
I.~Strelkovskaya, I.~Solovskaya, A.~Makoganiuk, and T.~Rodionova, ``Multimedia
  traffic prediction based on wavelet-and spline-extrapolation,'' in \emph{2020
  IEEE International Black Sea Conference on Communications and Networking
  (BlackSeaCom)}.\hskip 1em plus 0.5em minus 0.4em\relax IEEE, 2020, pp. 1--5.

\bibitem{l3}
Z.~Tian, ``Network traffic prediction method based on wavelet transform and
  multiple models fusion,'' \emph{International Journal of Communication
  Systems}, vol.~33, no.~11, p. e4415, 2020.

\bibitem{dt1}
Z.~Liu and L.~Menzel, ``Identifying long-term variations in vegetation and
  climatic variables and their scale-dependent relationships: A case study in
  southwest germany,'' \emph{Global and Planetary Change}, vol. 147, pp.
  54--66, 2016.

\bibitem{dt2}
S.~O. Mousavizadeh~Kashi and M.~Akbarzadeh, ``A framework for short-term
  traffic flow forecasting using the combination of wavelet transformation and
  artificial neural networks,'' \emph{Journal of Intelligent Transportation
  Systems}, vol.~23, no.~1, pp. 60--71, 2019.

\bibitem{scikit-learn}
F.~Pedregosa, G.~Varoquaux, A.~Gramfort, V.~Michel, B.~Thirion, O.~Grisel,
  M.~Blondel, P.~Prettenhofer, R.~Weiss, V.~Dubourg, J.~Vanderplas, A.~Passos,
  D.~Cournapeau, M.~Brucher, M.~Perrot, and E.~Duchesnay, ``Scikit-learn:
  Machine learning in {P}ython,'' \emph{Journal of Machine Learning Research},
  vol.~12, pp. 2825--2830, 2011.

\bibitem{lee2019pywavelets}
G.~Lee, R.~Gommers, F.~Waselewski, K.~Wohlfahrt, and A.~O'Leary, ``Pywavelets:
  A python package for wavelet analysis,'' \emph{Journal of Open Source
  Software}, vol.~4, no.~36, p. 1237, 2019.

\end{thebibliography}

\vspace{12pt}

\end{document}